\documentclass[pdflatex,sn-mathphys-num]{sn-jnl}% Math and Physical Sciences Numbered Reference Style
\usepackage{graphicx}
\usepackage{xcolor}

\usepackage{multirow}
\usepackage{amsmath,amssymb,amsfonts}
\usepackage{amsthm}
\usepackage{mathrsfs}%
\usepackage[title]{appendix}%
\usepackage{xcolor}%
\usepackage{textcomp}%
\usepackage{manyfoot}%
\usepackage{booktabs}%
\usepackage{algorithm}%
\usepackage{algorithmicx}%
\usepackage{algpseudocode}%
\usepackage{listings}%
\theoremstyle{thmstyleone}%
\newtheorem{theorem}{Theorem}%  meant for continuous numbers
\newtheorem{proposition}[theorem]{Proposition}% 

\theoremstyle{thmstyletwo}%

\theoremstyle{thmstylethree}%

\def\blue{\color{blue}}

\colorlet{blue}{black}

\def\bop{\noindent {\bf Proof.\ }}
\def\eop{\hfill $\Box$ \vspace{0.3 true cm}}

\def\card{{\rm card}}

\def\cH{{\mathcal H}}
\def\cX{{\mathcal X}}
\def\cB{{\mathcal B}}
\def\cE{{\mathcal E}}
\def\cG{{\mathcal G}}
\def\cP{{\mathcal P}}

\def\P{{\rm P}}
\def\E{{\rm E}}

\def\bR{{\mathbb R}}
\def\bN{{\mathbb N}}

\def\n{\noindent}

\def\blue{\textcolor{blue}}

\begin{document}

\title[Article Title]{Networks with Finite VC Dimension: \\ Pro and Contra}


\author*[1]{\fnm{V\v{e}ra} \sur{K\r{u}rkov\'{a}}}\email{vera@cs.cas.cz}

\author[2]{\fnm{Marcello} \sur{Sanguineti}}\email{marcello.sanguineti@unige.it}

\affil*[1]{\orgdiv{Institute of Computer Science}, \orgname{Czech Academy of Sciences}, \orgaddress{\street{Pod Vod\'{a}renskou v\v{e}\v{z}\'\i,  2}, \city{Prague}, \postcode{18207},  \country{Czech Republic}}}

\affil[2]{\orgdiv{Department of Computer Science, Bioengineering,
Robotics, and Systems Engineering (DIBRIS)}, \orgname{University of Genova}, \orgaddress{\street{Via Opera Pia, 13}, \city{Genova}, \postcode{16145}, \country{Italy}}}


\abstract{
\blue{Neural networks are expected both to learn from data and to approximate task-relevant functions with sufficient accuracy. While finite Vapnik-Chervonenkis (VC) dimension provides a classical guarantee of learnability through polynomial growth of the hypothesis class and uniform convergence of empirical errors, its implications for approximation accuracy have not been systematically examined. We develop a unifying probabilistic framework in which the relevance of functions to a given class of tasks is encoded by a probabilistic prior. Rather than studying worst-case bounds, we investigate how the distribution of approximation errors depends on the growth of sets of network input-output functions. Our results reveal a fundamental asymmetry: polynomial growth ensures uniform convergence of empirical errors, and hence learnability, but can cause approximation errors to concentrate around their mean that may be large. This mean is largest under a uniform prior, demonstrating the practical usefulness of uniform approximation theorems. It is shown that good approximation accuracy depends critically on the task distribution being aligned with the network's input-output function class. We further discuss the influence of network depth and the number of units on both the learnability and approximation accuracy of ReLU networks.}
}

%%================================%%


\maketitle

\section{Introduction}

\begingroup
\color{blue}

For the practical performance of neural networks, it is desirable not only that they {\em learn from samples of data}, but also that they {\em compute with sufficient accuracy} the tasks relevant to a given application domain. The classical guarantee of learnability \cite{vach71} is the finiteness of the Vapnik-Cervonenkis (VC) dimension of a hypothesis class. Finite VC dimension limits the growth function of the class  (e.g., the set of input-output functions realizable by a network), causing its growth to change from exponential to polynomial and thereby yielding uniform convergence of empirical errors. The approximation capabilities of neural networks, in contrast, have traditionally been studied through rates of decrease of approximation errors with growth of network complexity,  characterized by network width and/or depth. Typically, approximation accuracy is quantified in terms of worst-case errors over classes of functions defined by norm constraints that impose various forms of smoothness, such as Sobolev \cite{yazh20} and variational norms \cite{ba93,vk12}.

In this paper, we introduce a unified framework for studying learnability and approximation accuracy. As a {\em prior defining the classes of functions to be approximated}, we consider the relevance of functions, represented by probabilities that reflect their likelihood of occurrence within a given task. Rather than focusing on worst-case approximation errors, we study {\em distributions of approximation errors} and investigate how these distributions depend on network complexity, measured by the {\em growth function of the sets of input-output functions} realizable by the network.

Because contemporary neural networks are routinely applied to large datasets, we study learnability and function approximation in a high-dimensional setting. We show that the polynomial growth of the sets of input-output functions that guarantees learnability also makes these sets ``geometrically thin'' within the space of binary classifiers, which can, in turn, limit the approximation capacity of the network. 

To formalize our approach, we represent classifiers as vectors in spaces whose dimension is equal to the size of the dataset, which is typically very large, and exploit properties of high-dimensional geometry. In particular, we use the concentration-of-measure phenomenon \cite{misc86,le01,dupa09}, according to which sufficiently smooth functions of many random variables concentrate sharply around their mean values. We show that both the approximation-error and empirical-error functionals satisfy an appropriate smoothness condition, which can be interpreted as a coordinate-wise Lipschitz property. This allows us to apply McDiarmid's bounded-differences concentration inequality \cite{mc89}. Combining the resulting concentration bounds for the two error functionals with the polynomial growth of the sets of network input-output functions, we establish their almost-deterministic behavior in high dimensions.

For the concentration of approximation errors, it is sufficient to assume independence of the random variables induced by a function drawn from the probabilistic prior. For learnability, however, a stronger i.i.d. assumption is required: convergence of the empirical error to the theoretical error relies on all random variables having the same mean value.

Our results highlight a fundamental difference between the roles played by finite VC dimension in statistical learning theory and in neural-network approximation theory. For empirical errors, finite VC dimension ensures uniform convergence and hence supports learnability. For approximation errors, by contrast, it leads to concentration around potentially large values unless the network contains input-output functions that are strongly correlated with the underlying probability distribution. Thus, learnability and approximation accuracy impose distinct requirements, even though both are governed by the same growth properties of the underlying hypothesis class.

The resulting estimates of approximation errors show that the polynomial growth of the sets of input-output functions associated with networks of finite VC dimension is insufficient to compensate for the exponentially decreasing volume of their neighborhoods in high-dimensional spaces. In particular, in the absence of prior knowledge, when a uniform distribution over functions is a natural assumption, approximation errors concentrate around relatively large values. This reveals practical limitations on the expressive power of neural networks. Although the universal approximation property guarantees that sufficiently large networks can approximate arbitrary functions, achieving good accuracy for almost all functions requires a high VC dimension and, consequently, high model complexity. Conversely, our results show that when the data distribution is highly non-uniform, relatively small neural networks may achieve good approximation accuracy.

Our proofs rely on geometric properties of high-dimensional spaces, often referred to as the ``blessing of dimensionality"  \cite{do00,goty18,goal19,vk19a}. These include the exponential growth of quasi-orthogonal dimension \cite{kavk93}, concentration of volume and surface measure in high-dimensional spheres \cite{ba97}, and dimensionality reduction via random projections  \cite{joli84}. Related applications of these phenomena in neurocomputing address highly variable functions  \cite{vksa16}, robustness, and the intrinsic dimensionality of data  \cite{sual23,baal23}. Applying these properties  to both the empirical error and approximation error functional implies concentration of empirical errors around the theoretical error and hence almost deterministic behavior of approximation errors.

The dependence of learnability and the distribution of approximation errors on network architecture is indirect, mediated by the effect of architecture on the network's growth function. We investigate this dependence for the currently popular class of deep ReLU networks, using recent estimates of their VC dimension \cite{baal19,baal21}. Some preliminary results were presented in the conference proceedings \cite{vk23}.

\endgroup

The paper is organized as follows. Section \ref{sec:prel} introduces the basic concepts and notation. Section \ref{sec:prob} develops a probabilistic framework for function approximation and establishes smoothness conditions that guarantee concentration of both approximation and empirical errors. Section \ref{sec:conc} presents the main results on accuracy and consistency. Section \ref{sec:VC} combines probabilistic bounds with estimates of growth functions for deep ReLU networks and derives implications for their suitability in tasks governed by probability distributions. Finally, Section \ref{sec:disc} summarizes and discusses the main findings.

 \section{Preliminaries}
 \label{sec:prel}

We use the \blue{following}  notation: $\bN_{+}$ denotes the set of {\em positive integers}, $(\bR^m, \| . \|_2)$  the {\em $m$-dimensional Euclidean space with the $l_2$-norm}.
The {\em distance of a function $f$} in a normed linear space $(\cX, \|.\|_{\cX})$ from its subset $\cH$ is denoted
$$\| f - \cH\|_{\cX} := \inf_{h \in \cH} \|f - h\|_{\cX}.$$
By $f^{\circ} := \frac{f}{\| f \|_{\cX}}$  we denote the {\em normalized function}.
 For $U \subseteq \bR^d$, we denote
$$\cB(U) : = \{ f : U \to \{-1,1\} \}.$$
The set $\cB(U)$ models binary classifiers on $U$. We investigate
approximation and learning of classifiers on {\em finite data sets}.
So we consider functions from $\cB(X)$  where $$X= \{x_1, \dots, x_m\} \subset \bR^d.$$ 
\n Typically $m>>d$. The set $\cB(X)$ can be identified with the {\em Hamming cube} $\{-1,1\}^m$ and its elements with $m$-dimensional vectors. For a set $\cH \subset \cB(\bR^d)$ we denote
$$\cH(X):= \cH_{|_X} = \{ h|_{X}: X \to \{-1,1\} \, | \, h \in \cH\} \subset \cB(X)\,.$$

{\em Feedforward neural networks} compute parameterized families of input-output (I/O) functions.  They are determined by network architectures described by directed acyclic graphs $\cG$, where nodes represent computational units and edges represent connections between them. In practical applications, network inputs belong to large finite sets, so we investigate networks with finite sets of inputs $X =\{x_1, \dots, x_m\}  \subset \bR^d$ representing input data in the form of vectors of features. We focus on networks with single binary-valued outputs from $\{-1,1\}$.
Sets of I/O functions of such networks are subsets of $\cB(X)$ formed by vectors having $l_2$-norms equal to $\sqrt{m}$.

In {\em multilayer networks}, units are arranged in layers, network inputs are viewed as layer $0$, outputs as layer $L$,  the layers $l=1, \dots, L-1$ are called {\em hidden layers}, and $L$ is called the {\em network depth}.
A biologically inspired computational unit called {\em perceptron}  applies a fixed {\em activation function}  $\psi: \bR \to \bR$ to an affine function with varying parameters. A perceptron computes functions of the form
$$\psi(v \cdot . +b): \bR^d \to \bR,$$
\n where $v \in \bR^d$ is called a {\em weight vector}, $b \in \bR$ a {\em bias}, and $v \cdot x = \sum_{i=1}^d v_i x_i$ is the {\em scalar product} of the weight vector with input vector $x$ (weighted sum of inputs). Originally, perceptrons were endowed with sigmoidal activation functions representing hard or soft thresholds, such as the {\em Heaviside} function defined as  $\theta(t) = 0$ for $t\ \le 0$ and $\theta(t) =1$ for $t \geq 1$. Currently, the most widespread computational units are {\em rectified linear units} (ReLU), which are perceptrons with the activation function $\rho(t) = \max(0,t)$.

The parameters of computational units and their connections are optimized during learning.
A {\em loss function} $\ell: X \times Y \to \bR_+$  measures the difference between the desired value $y \in Y$ and the output $h(x)$ computed by the I/O function $h$.

\section{Probabilistic framework for learnability and function approximation}
\label{sec:prob}

We study approximation and learning of binary-valued functions on finite sets of data $X = \{x_1, \dots, x_m\} \subset \bR^d$ that are constrained by a probabilistic prior that models the likelihood of a function occurring in a given type of application. We  assume there is a {\em probability distribution} ${\P}$ on $\cB(X)$, according to which functions to be computed are chosen. 
We consider the case when $\P$ can be expressed as the product
$$\P(f) := \prod_{i=1}^m \P_i (f(x_i)) =  \prod_{i=1}^m \P(f(x_i)|x_i),$$
\n which implies that the random variables $f(x_1), \dots, f(x_m)$ are {\em independent}. Unless specified otherwise, we do not assume that they are identically distributed.

\blue{Further, we assume that in addition to the probability distribution $\P$ on $\cB(X)$, there is a probability $\P_X$ on $X$. The {\em training samples} of $n$ pairs of input-output data  (where $n$ is typically much smaller than $m$)} are drawn from the probability $\cP_n$ defined in terms of both  $\P$ and $\P_X$ as

$$\cP_n(S) := \prod_{j=1}^n \P_X(x_{i_j})\P(y_{i_j}| x_{i_j}).$$

\n For any hypothesis set $\cH(X)$ and a {\em loss function} 
$\ell: X \times \{-1,1\} \to \bR_+$ and $n \in\bN_+,$
 we denote by  $\cE_{\ell,h,n}: (X \times \{-1,1\})^n \to \bR_+$ the {\em empirical error functional}
defined for a sample $S = ( (x_{i_1},y_{i_1}), \dots, (x_{i_n}, y_{i_n}) )$ as
 $$\cE_{\ell,h,n}(S) := \frac{1}{n} \sum_{j=1}^n \ell \bigl ( y_{i_j},h(x_{i_j}) \bigr ).$$

\n We denote by $\eta_{h}: \cB(X) \to \bR_+$  the {\em square of the $l_2$-distance of $h^{\circ}$ from normalized functions from $\cB(X)$}
 $$\eta_{h}(f) := \| f^{\circ} - h^{\circ}\|_2^2= \frac{1}{m} \| f -h\|_2^2.$$

 \n We study $\eta_{h}$ as a function of random variables $f(x_1), \dots, f(x_m)$ drawn from $\{-1,1\}$ according to ${\P(f)}$ and the functional $\cE_{\ell,n,h}$  as a function  of random variables $(x_{i_1}, y_{i_1}), \dots, (x_{i_n}, y_{i_n})$ drawn from $(X \times \{-1,1\})$ according to ${\cP}_n(S)$.

We employ a concentration inequality that holds for functions of independent random variables satisfying a smoothness assumption, which can be viewed as a coordinate-wise Lipschitz property.
A function
$\phi: A_1 \times \ldots \times A_m \to \bR$ satisfies the {\em bounded differences (BD) condition
with the parameter vector} $c= (c_1, \dots, c_m)$ if for all $i=1, \dots, m$ and all vectors $a, a' \in  A_1 \times \ldots \times A_m$ that differ only in the $i$-th coordinate,
 \begin{equation}
 | \phi(a) - \phi(a')| \leq c_i.
\label{eq:BD}
\end{equation}

\n Note that with sufficiently large parameters, the BD condition holds for every bounded function. The following theorem by McDiarmid \cite[p.70]{dupa09},\cite{mc89} shows that when the coefficients are small enough, then the BD condition implies concentration of function values around their mean value.

\begin{theorem}[McDiarmid Bound]
Let $A_1, \dots, A_m \subset \bR^d$, $\phi: \prod_{i=1}^m A_i  \to \bR$ satisfies the bounded differences condition with the vector of parameters $c:= (c_1, \ldots, c_m)$,  $\cP = \prod_{i=1}^m \cP_i$ a probability on $\prod_{i=1}^m A_i$, and $z_1, \ldots, z_m$ be independent random variables with values in ranges $A_1, \ldots, A_m$, drawn according to $\cP_i$, resp. Then for every $\lambda>0$,

\begin{equation}
\cP\Bigl [ \, \Bigl | \phi(z_1, \ldots, z_m) - \E(\phi)\Bigr | > \lambda \Bigr ]  \leq e^{-2\lambda^2/\| c \|_2^2}\,.
\end{equation}
\label{th:MD}
\end{theorem}

\n When $\|c\|_2^2 \leq \frac{s}{,m}$  for $c$ constant, and $\lambda$ is fixed, then  the probability that values of $\phi$ are close to  $\E(\phi)$  converges with  $m$ increasing  quickly to $1$. Thus, the behavior of $\phi$ is almost deterministic.

To apply the McDiarmid bound to the functionals $\eta_{h}$  and $\cE_{\ell,h,n}$, we first explore parameters for which they satisfy the BD condition.

\begin{proposition}
Let $X = \{x_1, \dots, x_m\} \subset \bR^d$,  $h \in \cB(X)$, $\ell: \{-1,1\} \times \{-1,1\} \to [0,b]$ be a loss function.

 (i) $\eta_{h}: \cB(X) \to \bR_+$ satisfies the BD condition with the parameter vector $c$ with all $c_i= \frac{4}{m}$;

(ii) $\cE_{\ell,h,n}: (X \times \{-1,1\})^n \to \bR_+$ satisfies the BD condition with the parameter vector $c$ with all $c_i=\frac{2b}{n}$.
\label{pr:BD}
\end{proposition}

\bop
Without loss of generality, we assume that in both cases the coordinate to be changed is the first one.

(i) Let $f(x_1) = y_1$ and $f'(x_1)= {y_1}'$ and $f(x_i) = f'(x_i) = y_i$ , $i=2, \dots, m$. Then we have\\
$\frac{1}{m}  \Bigl |
\sum_{i=1}^m \bigl ( f(x_i) - h(x_i) \bigr )^2  -
\sum_{i=1}^m \bigl ( f'(x_i) - h(x_i) \bigr )^2
\Bigr | =
\frac{1}{m} \Bigl |
\bigl (y_1 - h(x_1)\bigr )^2 - \bigl ({y_1}' - h(x_1) \bigr )^2   \Bigr |  \leq \frac{4}{m}$.

(ii)
$ \frac{1}{n} \Bigl |\,  \ell(y_1, h(x_1)) + \sum_{i=2}^n  \ell(y_i,h(x_i)) - \ell(y'_1, h(x'_1)) - \sum_{i=2}^n  \ell(y_i,h(x_i)) \, \Bigr | =$\\
$ \frac{1}{n} \Bigl | \ell(y_1, h(x_1)) - \ell(y'_1, h(x'_1)) \Bigr |  \leq \frac{2b}{n}.$
\eop

\n Proposition \ref{pr:BD} shows that for every $h \in \cB(X)$, the functional $\eta_h$ satisfies the BD condition with $\| c\|^2_2= \frac{16}{m}$ and the functional $\cE_{\ell,h,n}$ with $\| c\|^2_2= \frac{4b^2}{n}$.
So  for a sufficiently large $m$ and $n$, resp., these two functionals behave almost deterministically.

\section{Accuracy of approximation and learnability}
\label{sec:conc}

 On infinite domains ($\bR^d$ or its open subsets), I/O functions of neural networks of many types determine network parameters {\em uniquely up to permutations and sign flips} (see, e.g., \cite{su92,also93,vkka94,vkka96}).
On finite domains, neural network function representations are typically highly redundant. In particular, in the case of families of binary-valued functions, infinitely many parameterizations generate only finitely many functions.  Vapnik and Chervonenkis \cite{vach71} discovered the crucial role of these sizes in learnability from data samples. 
 For a family of binary-valued functions $\cH$
on any $U \subset \bR^d$,  the {\em growth function} $\Pi_{\cH}(m): \bN_{+} \to \bN_{+}$ is defined  as
$$ \Pi_{\cH}(m) := \max_{V \subset U, \card V =m} \card \cH _{| \, V}.$$

The next theorem proves that the growth function $\Pi_{\cH}$ also influences 
the concentration of errors in the approximation of functions by $\cH$. For each $h \in \cH(X)$, let
$$\mu_h= \E(\eta_h)$$
\n denote  the {\em mean value} of $\eta_h$ with respect to $\P$ and
$$\mu_{\cH(X)}= \min_{h \in \cH(X)} \mu_h.$$

\begin{theorem}
Let $\cH \subset \cB(\bR^d)$, $X=\{x_1, \dots, x_m\} \subset \bR^d$, $\P$ be a product probability distribution on $\cB(X)$ and $\cH(X) \subset \cB(X)$. Then for every $\lambda>0$\\
$$ (i)  \quad \P \Bigl [ \mu_{\cH(X)} -\lambda \leq \|f^{\circ} - \cH(X)^{\circ} \|^2_2 \Bigr ] > 1 - \Pi_{\cH}(m) \, e^{-\frac{m\lambda^2}{8}};$$
$$ (ii)  \quad \P \Bigl [ \|f^{\circ} - \cH(X)^{\circ} \|^2_2 \leq \mu_{\cH(X)} + \lambda \Bigr ] > 1 - e^{-\frac{m\lambda^2}{8}}.$$
\label{th:l2H}
\end{theorem}

\bop
Combining the McDiarmid bound (Theorem \ref{th:MD}) with Proposition \ref{pr:BD} we obtain for every $h \in \cH$,
\begin{equation}
\P \Bigl [ \mu_h -\lambda\leq  \| f^{\circ} - h^{\circ} \|^2_2
\leq  \mu_h + \lambda \Bigr ] > 1- e^{-\frac{m\lambda^2}{8}}.
\label{eq:h}
\end{equation}
Hence $$\P \Bigl [ ( \forall h \in \cH(X))\, \Bigl ( \mu_h -\lambda\leq  \| f^{\circ} - h^{\circ} \|^2_2 \Bigr )   \Bigr ]
> 1- \card \cH(X) \, e^{-\frac{m\lambda^2}{8}} \geq 1 - \Pi_{\cH}(m) \, e^{-\frac{m\lambda^2}{8}}.$$
\n As $\mu_{\cH(X)} \leq \mu_h$ for all $h \in \cH(X)$,  (i) holds.
As $\cH(X)$ is finite, for every $f \in \cB(X)$ there exists  $h_f \in \cH(X)$ for which $\| f - \cH(X)\|_2 = \| f - h_f\|$.
So (ii) follows.
\eop

\begin{figure}[ht]
\centering
\includegraphics[width=12cm]{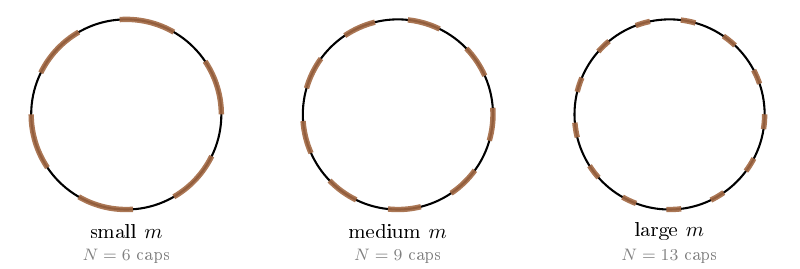}
\caption{\blue{Illustration of Theorem \ref{th:l2H} (i). $N=\Pi_H(m)$ caps scattered on $S^{m-1}$. As $m$ grows, each cap shrinks exponentionally while $N$ grows only polynomially.}}
\label{fig:capsC}
\end{figure}

To obtain conditions on the concentration of empirical errors applying McDiarmid's bound, we have to strengthen assumptions. We assume that the probabilities $\P_1, \dots, \P_m$ on $\{-1,1\}$ are not only independent but also identically distributed; \blue{otherwise the mean values would be sample-specific} . We set $\P_0 = \P_i$ for all $i=1, \dots, m$, and denote by
$$\cE_{\ell}(h) = \E(\ell(y,h(x))$$
\n the mean value of the functional $\ell(y, h(x))$ with respect to the probability 
$${\cP} =\P_X(x)\P_0(y|x).$$

\begin{theorem}
Let $X=\{x_1, \dots, x_m\} \subset \bR^d$, $\P$ be a probability distribution on $\cB(X)$ such that  $\P= \prod_{i=1}^m \P_i$ where $\P_i$ are i.i.d.,
$\cH \subset \cB(\bR^d)$, and $\ell: \{-1,1\}  \times \{-1,1\} \to [0,b]$ be a loss function. Then for every $h \in \cH(X)$, every positive integer $n \leq m$, every random sample $S=(x_{i_1}, y_{i_1}), \dots, (x_{i_n}, y_{i_n})$, and every $\lambda >0$
$${\cP_n} \bigl [ \, \bigl ( \forall h \in \cH(X) \bigr ) \left | \, \cE_{\ell,h,n,}(S) - \cE_{\ell}(h) \, \bigr  |  \leq \lambda \, \right ]\,  > 1-  \Pi_{\cH}(n) e^{-\frac{n \lambda^2}{2b^2}}.$$
\label{th:empH}
\end{theorem}

\bop
Let $U_n \subset X$ with $\card \, U_n =n$ and
$S = ( \, (x_{i_1},y_{i_1}), \dots, (x_{i_n}, y_{i_n}) \, )$ be a sample such that $\{x_{i_1}, \dots, x_{i_n}\}=U_n$. As the random variables $(x_{i_j},y_{i_j})$ are i.i.d., for every $h\in \cB(U_n)$,
we have
$$\E \bigl (  \cE_{\ell, h,n}(h) \bigr ) = \E \bigl( \frac{1}{n}
\sum_{j=1}^n \ell(y_{j_i}, h(x_{j_i})) \, \bigr )  =
 \E \bigl (\ell(y,h(x) \bigr ) = \cE_{\ell}(h),$$
 \n where the mean value is considered with respect to $\cP= \P_X(x)\P_0(y|x)$.
By Theorem \ref{th:MD},
$$ \cP \bigl [ \,  \left | \, \cE_{\ell,h,n}(S) - \cE_{\ell}(h) \, \bigr  |  > \lambda \, \right ]\, \leq e^{-\frac{n \lambda^2}{2b^2}}.$$

\n  As $\cP_n (S)=   \prod_{j=1}^n \P_X(x_{i_j}) \Bigl ( \P_0(y_{i_j} | x_{i_j}) \Bigr )^n$, we have
$$ \cP_n \bigl [ \, (\forall h \in \cH)  \left | \, \cE_{\ell,h,n}(S) - \cE_{\ell}(h) \, \bigr  |  > \lambda \, \right ]\,  \leq \Pi_{\cH}(n) e^{-\frac{n \lambda^2}{2b^2}}.$$
\eop

\n For $\lambda = n^{-1/4}$, Theorems \ref{th:l2H} and \ref{th:empH} imply
 \begin{equation}
 \P \Bigl [ \mu_{\cH(X)} - n^{-1/4} \leq \|f^{\circ} - \cH(X)^{\circ} \|^2_2 \Bigr ] > 1 - \Pi_{\cH}(n) \, e^{-\frac{n^{1/2}}{8}}
\label{eq:l2Hi}
\end{equation}

So
 \begin{equation}
 \P \Bigl [  \bigl \| f^{\circ} - \cH(X)^{\circ} \bigr \|^2_2 \leq \mu_{\cH(X)} + n^{-1/4} \Bigr ] > 1 - e^{-\frac{n^{1/2}}{8}}.
\label{eq:l2Hii}
\end{equation}

\begin{equation}
{ \cP}_n \bigl [ \bigl ( \forall h \in \cH(X) \bigr ) \,  \left | \, \cE_{\ell,h,n}(S) - \cE_{\ell}(h) \, \bigr  |  \leq n^{-1/4} \, \right ]\,  > 1-  \Pi_{\cH}(n) e^{-\frac{n^{1/2}}{2b^2}}.
\label{eq:empH}
\end{equation}
\eop

Theorem \ref{th:l2H} shows that the size of the set $\cH(X)$ strongly influences the concentration of approximation errors around  their mean value. When the growth function $\Pi_{\cH}(m)$ does not outweigh the term $e^{- \frac{m \lambda^2}{8}}$, the approximation errors with respect to $\cH$ behave almost deterministically. Similarly, Theorem \ref{th:empH} shows that when the growth function $\Pi_{\cH}(n)$ does not outweigh
 $e^{-\frac{n \lambda^2}{2b^4}}$, then for all functions $h \in \cH(X)$  and for almost all samples of size $n$, the empirical errors $\cE_{\ell,h,n}$ are close to the theoretical error $\cE_{\ell}(h)$. 

The convergence of empirical errors implied by Theorem \ref{th:empH} is a well-known result \cite{vach71} in statistical learning theory concerning the consistency of learning from data samples. \blue{Our contribution is 
pairing this theorem in a unifying framework with Theorem \ref{th:l2H}. Both theorems are based on the same concentration property, but applied to different functionals. The i.i.d. condition in Theorem \ref{th:empH} is needed to guarantee that the mean value is not sample-specific.}

\begingroup
\color{blue}

In the absence of prior knowledge, we have to assume that all functions in $\cB(X)$ are equally likely to occur in a task, i.e., that $\P$ is uniform. By symmetry, this implies $\E(\eta_h)=2$ for all $h \in \cB(X)$, and hence $\mu_{\cH(X)}=2$ for any class $\cH(X)$. Consequently, for any network class with  growth function $\Pi_{\cH(X)}$ increasing with $\card X = m$ slowler than $e^{\frac{m\lambda^2}{8}}$,  approximation errors of uniformly chosen functions concentrate around the mean value $\mu_{\cH(X)}=2$. This finding shows  a practical limitation of the universal approximation property: unless the size of the function class exceeds 
$e^{\frac{m{\lambda}^2}{8}}$, the majority of functions cannot be approximated with errors smaller than $2 - \lambda$ \blue{(see Figure \ref{fig:one_capC})}. In particular, for $\lambda =  m^{-1/4}$, almost all functions incur approximation errors of at least $2 - m^{-1/4}$.

\begin{figure}[ht]
\centering
\includegraphics[width=6cm]{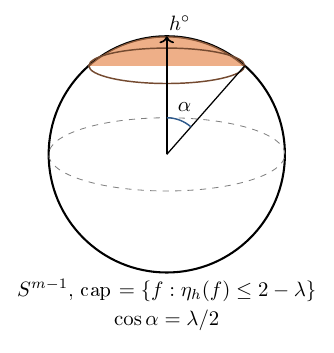}
\caption{\blue{Classifiers within $\lambda$ of $h$ in the case of uniform probability.}}
\label{fig:one_capC}
\end{figure}

When the distribution $\P$ is non-uniform, the situation can be markedly different. It may happen that
$\mu_{\cH(X)} = \min_{h \in \cH(X)} \E(\eta_h)$ is smaller or equal to the  desired accuracy. 
In this case, there exists a function $h^* \in \cH(X)$ such that $\E(\eta_{h^*})$ is sufficiently small. Inspection of  the proof of Theorem \ref{th:l2H} shows that approximation errors for almost all functions concentrate around $\E(\eta_{h^*}) = \E | f^{\circ} - h^{*\circ}|_2^2$. Therefore, with respect to $\P$, most functions can be well approximated by this single function $h^*$. The cases of uniform and non-uniform probability are illustrated in Figure \ref{fig:unif_VS_non-unifC}.

\begin{figure}[ht]
\centering
\includegraphics[width=13.5cm]{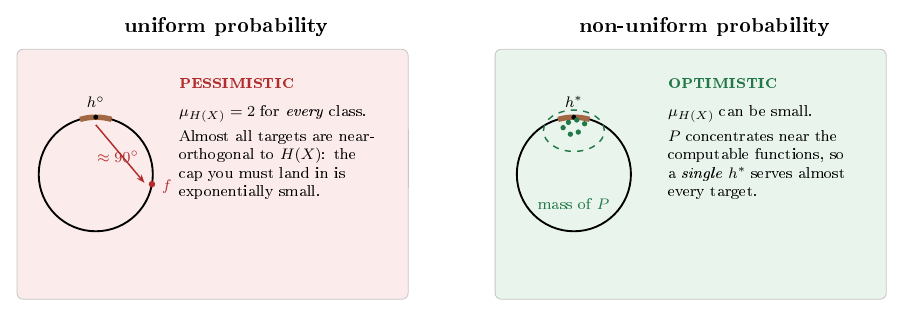}
\caption{\blue{The cases of uniform and non-uniform probability.}}
\label{fig:unif_VS_non-unifC}
\end{figure}

\endgroup

\section{Neural networks with finite VC dimension}
\label{sec:VC}

The theorems in the previous section imply that both the concentration of approximation errors and the convergence of empirical errors hold for networks whose sets of I/O functions grow polynomially with the size $m$ of their domain.. 
\blue{A fundamental result by Vapnik and Chervonenkis \cite{vach71} establishes that the growth function of any set $\cH$ is either equal to  $2^m$ for all m, or there exists a finite threshold, called the {\em VC dimension}, beyond which the growth function is bounded by a polynomial in $m$.}
We denote
$$d_{V}(\cH) = \max_{m \in \bN} \{\Pi_{\cH}(m) = 2^m\}, $$
\n if such maximum exists, otherwise $d_{V}(\cH) = \infty$.
Vapnik and Chervonenkis \cite[Theorem 1]{vach71} proved that for $m>d_V$
\begin{equation}
\Pi_{\cH}(m) \leq m^{d_V +1}.
\label{eq:VC}
\end{equation}

\n A slightly better result was obtained independently by Shelah \cite{sh72} and Sauer \cite{sa72}.

As any class with finite VC dimension has a growth function bounded by a polynomial, Theorem \ref{th:l2H} implies that approximation errors of networks whose I/O function classes have finite VC dimension concentrate around their mean values. Moreover, the smaller $\Pi_{\cH}(m)$, the higher the probability that the approximation accuracy for a randomly drawn function is close to $\mu_{\cH(X)}$. 

While consistency in learning from data is a desirable property of a class of neural networks, the concentration of approximation errors near $\mu_{\cH(X)}$ may impair approximation capability.  If $\mu_{\cH(X)}$ is not sufficiently small, then almost all functions drawn according to $\P$ cannot be approximated to a desired accuracy by  networks with sets of I/O functions of finite VC dimension. In this case, their squared normalized $l_2$ errors are at least $\mu_{\cH(X)} - \lambda$.
On the other hand, if $\mu_{\cH(X)}$ is sufficiently small, then by the upper bound in Theorem \ref{th:l2H}(ii),
 almost all functions can be approximated with error at most $\mu_{\cH(X)} + \lambda$ using a single function
 $h^*$ (i.e., any function for which $\mu_{\cH(X)} = \mu_{h^*}$). Thus, network classes with finite VC dimension are not capable of approximating most functions in $\cB(X)$ with high accuracy. Interpreting functions as vectors in $\{-1,1\}^m$, this reflects the fact that, in high dimensions, most functions are nearly orthogonal to any hypothesis class $\cH(X)$ whose growth function is only polynomial in $m$.

The problem of estimating the number of binary-valued functions realizable on $m$-point subsets of $\bR^d$ has been studied since the 19th century. Schl\"{a}fli \cite{sc01,sc50} derived an upper bound on the number of linearly separable dichotomies of $m$ points in $\bR^d$. In the language of neural networks, this corresponds to the maximal number of functions that can be implemented by a single perceptron with a Heaviside activation function on such subsets. Later, Cover \cite{co65} established function-counting theorems that estimate the number of dichotomies induced by certain nonlinear decision surfaces, including hyperspheres and hypercones.

Bartlett et al. \cite{baal98,baal19} studied the VC dimension of deep networks composed of perceptrons with piecewise linear activation functions. An inspection of \cite[Theorem 7 and Remark 9]{baal19} shows that, if $ReLU(L,W)$ denotes the class of input-output functions implemented by ReLU networks with $L$ layers and a total of $W$ parameters, then
$$\Pi_{ReLU(L,W)}(m) \leq \bigl (  4eLm \bigr )^{LW}.$$

\begingroup
\color{blue}

\n Thus, the growth function of deep ReLU networks depends polynomially on the sample size $m$, with degree equal to the product $LW$.  For $X=\{x_1, \dots, x_m\} \subset \bR^d$, a product probability distribution $\P$ and  $\lambda>0$, Theorem \ref{th:l2H} implies the bound

\begin{equation}
 \quad \P \Bigl [ \mu_{ReLU(L,W)(X)} -\lambda \leq \|f^{\circ} - \cH(X)^{\circ} \|^2_2 \Bigr ] > 1 - \bigl (  4eLm \bigr )^{LW} \, e^{-\frac{m\lambda^2}{8}}\,
\label{eq:approx}
\end{equation}

\n and Theorem \ref{th:empH}  the bound

\begin{equation}
{\cP_n} \bigl [ \, \bigl ( \forall h \in \cH(X) \bigr ) \left | \, \cE_{\ell,h,n}(S) - \cE_{\ell}(h) \, \bigr  |  \leq \lambda \, \right ]\,  > 1-  \bigl (  4eLn \bigr )^{LW} e^{-\frac{n \lambda^2}{2b^2}}\,\,\,\,\, \forall n \leq m.
\label{eq:learn}
\end{equation}

\endgroup

In many practical settings, it is reasonable to assume that both $L$ and $W$ are much smaller than the cardinality $m$ of the domain $X$ (since networks with as many layers or parameters as data points would be inefficient). Consequently, for sufficiently large $m$, the approximation errors of random functions by commonly used deep ReLU networks become highly concentrated and therefore behave almost deterministically. \blue{Almost all errors are close to  $\mu_{\cH(X)}= \mu_{h^*}$. So the network either approximates well almost all relevant functions or almost all functions are far from its set of I/O functions.}

\blue{Increasing network depth and width may improve approximation accuracy by reducing 
$\mu_{\cH(X)}$. However, by the bound (\ref{eq:learn}), it may also require a larger training sample size $n$ to ensure that the exponential term $e^{-\frac{n \lambda^2}{2b^2}}$ dominates the polynomial factor $\bigl( 4eLn \bigr)^{LW}$. }

\blue{Currently, some highly complex deep networks are exploited in applications. Such networks have very large VC dimensions, and so the concentration phenomenon may no longer occur. In such cases, the uniform convergence of empirical errors is not guaranteed. In \cite{baal21}, it was conjectured that such deep models can be decomposed into a high-complexity component that allows for a very good fit to the data, and a low-complexity component for which classical uniform convergence holds. This conjecture appears to be consistent with our results, which guarantee for the low-complexity component  classical uniform convergence of empirical errors, together with concentration of approximation errors around a single value, which might be improved in the high-complexity component.}

\section{Discussion}
\label{sec:disc}

\begingroup
\color{blue}

Our results reveal a fundamental issue: finite VC dimension of a network's
input-output (I/O) function class guarantees learnability through uniform
convergence of empirical errors, yet it simultaneously may impair approximation
accuracy. The polynomial growth of the set of input-output functions, which is the source of
learnability, also makes the class geometrically sparse in the
high-dimensional space of binary classifiers. Consequently, approximation
errors concentrate around the minimum mean distance $\mu_{\mathcal{H}(X)}$,
becoming almost deterministic as the dataset size $m$ increases. When
$\mu_{\mathcal{H}(X)}$ is large, almost all functions drawn from the task
distribution cannot be accurately approximated, regardless of how the network
is trained.

The behavior is sharply distribution-dependent. Under a uniform prior, nearly
all functions incur large approximation errors, since no single I/O function
lies close to most of the classifier space. Under a non-uniform distribution
concentrated near the function class, however, $\mu_{\mathcal{H}(X)}$ can be
small, and a single function $h^*$ suffices to approximate almost every
task-relevant function accurately (see Figure \ref{fig:unif_VS_non-unifC}). This shows that the task
distribution is as decisive as network architecture. By contrast, convergence of empirical errors
is distribution-independent, though it additionally requires i.i.d. data,
whereas our concentration bounds for approximation errors require only
independence.

For deep ReLU networks with $L$ layers and $W$ parameters, the growth
function satisfies $\Pi_{\mathcal{H}}(m)\leq(4eLm)^{LW}$. Larger depth and
width can reduce $\mu_{\mathcal{H}(X)}$ for some distributions, but do
not escape the fundamental issue: approximation errors remain concentrated,
and good performance ultimately depends on aligning the function class with
the task distribution rather than on model size.

Our findings show that learnability and approximation accuracy, although being controlled by the same growth properties of the underlying hypothesis class, impose different requirements. For empirical errors, finite VC dimension ensures uniform convergence and hence supports learnability. For approximation errors, by contrast, it leads to concentration around potentially large values unless the network contains input-output functions that are strongly correlated with the underlying probability distribution. 

Our results provide some insight into the recent discussion from \cite[p. 90]{baal21}, concerning near-optimal fits of very large deep networks to noisy training data together with excellent generalization performance.  In  \cite{baal21}, it was conjectured that such models can be decomposed into low-complexity ones, for which uniform convergence of empirical risk works, and high-complexity ones that enable good fit to data.  Our analysis provides support to the conjecture by demonstrating the fundamental trade-off between learnability and accuracy. A combination of low- and high-complexity models may be a way to achieve both objectives. 

Note that the large VC-dimension of deep ReLU networks is an anomaly with respect to classical machine-learning approaches: for a fixed number of parameters it increases with the number of layers exponentially, and so it becomes very large for commonly-used deep networks, which have hundreds of layers. In addition, an arbitrarily small perturbation of the ReLU nonlinearity may make the VC-dimension of the corresponding class change from a small constant to infinity \cite[Example 2.6]{baal21}, and the same phenomenon holds for a slightly perturbed sigmoid nonlinearity \cite[Lemma 7.2]{anba99}. 

The framework of this paper covers binary-valued functions on finite domains with the normalized $\ell_2$ metric. Extensions to real-valued outputs and other loss functions -- relevant, e.g.,
to neural approximation of differential and integral equations~(see \cite{paal25,paal26} and the references therein) --
and to correlated data~\cite{vksa21} are directions for future work.

\endgroup

\section{Acknowledgments}
V.K. was partially supported by the Czech Science Foundation grant GA25-15490S and the institutional support of the Institute of Computer Science RVO 67985807.
M.S. is a member of
GNAMPA-INdAM and Guest Scholar at IMT - School for Advanced Studies, Lucca,
Italy. He was supported by the projects PDGP
DIT.AD021.104 of INM-CNR (National Research Council of Italy),
where he is Research Associate, by project F/310027/02/X56 of the Italian Ministry of Enterprises and Made in Italy, and by the FISA-2022-00827 project UAV-FIRE  PRIN 2022S8XSMY of the Italian Ministry of Research.

\bibliography{vcdim}

\end{document}